\documentclass[final,1p,times]{elsarticle}

\usepackage{amssymb}

\usepackage{float}
\usepackage{varioref}
\usepackage{amsmath}
\usepackage[colorlinks]{hyperref}
\usepackage[noabbrev]{cleveref} 

\usepackage{lineno}
\usepackage{algorithm}
\usepackage[noend]{algpseudocode}
\usepackage{graphicx}
\usepackage{subcaption}
\usepackage{adjustbox}

\usepackage{subcaption}
\usepackage{multirow}

\usepackage{booktabs}

\journal{}

\begin{document}

\begin{frontmatter}

\title{Livestock Fish Larvae Counting using DETR and YOLO based Deep Networks}

\address[label1]{Universidade Católica Dom Bosco, Campo Grande, MS, Brazil}
\address[label2]{Agropeixe Ltda.}
\address[label3]{Universidade Federal de Mato Grosso do Sul, Campo Grande, MS, Brazil}


\author[label1]{Daniel Ortega de Carvalho}
\ead{daniel.ortega.carvalho@gmail.com}

\author[label1]{Luiz Felipe Teodoro Monteiro}
\ead{luiz.felipe.monteiro21@gmail.com}

\author[label2]{Fernanda Marques Bazilio}
\ead{fermarques.cbio@gmail.com}

\author[label1]{Gabriel Toshio Hirokawa Higa}
\ead{ra867467@ucdb.br}

\author[label1,label3]{Hemerson Pistori}
\ead{pistori@ucdb.br}

\begin{abstract}
Counting fish larvae is an important, yet demanding and time consuming, task in aquaculture. In order to address this problem, in this work, we evaluate four neural network architectures, including convolutional neural networks and transformers, in different sizes, in the task of fish larvae counting. For the evaluation, we present a new annotated image dataset with less data collection requirements than preceding works, with images of spotted sorubim and dourado larvae. By using image tiling techniques, we achieve a MAPE of 4.46\% ($\pm 4.70$) with an extra large real time detection transformer, and 4.71\% ($\pm 4.98$) with a medium-sized YOLOv8.
\end{abstract}


\end{frontmatter}

\section{Introduction}
\label{introducao}
Counting fish larvae is an important task in aquaculture, allowing for monitoring growth and development of the fish from early stages. Traditionally, fingerling identification is performed manually by specialists, which means that it is usually a slow and error-prone process that can compromise the development of an animal. The advancement of image processing technologies and also of machine learning have brought about new approaches for automating larvae counting, which may lead to greater agility in the processes of aquaculture~\cite{zhao2021application}.

The use of automation is increasing in the aquaculture industry, which is largely due to advancements in information technology. By using IT, it is possible to enhance monitoring of the fish and of their habitats, and even to manipulate them~\cite{barbedo2022review}. Among the technologies in use, artificial intelligence and, specifically, machine learning algorithms have become prominent because of their capacity of solving complex problems, such as those found in aquaculture.

As stated, fish larvae counting is an important yet difficult task. By using machine learning techniques for image processing, it may be possible to develop high performance models to identify and count larvae automatically. If possible, models that satisfactorily count fish larvae may become a tool to reduce the time required by the task, as well as its costs, leading to a more efficient and sustainable production~\cite{gonccalves2022using}.

In this work, we evaluate State-of-the-Art deep neural networks for object detection, including transformer-based architectures, used for counting larvae in high resolution images, some of them containing high density of individuals, along with a tiling technique. The smallest MAPE achieved in this work was 4.46 ($\pm 4.70$) with an extra large real-time detection transformer. Differently from other works, such as that by \citet{costa2023counting}, we use data that adequately represents real-life scenarios, instead of increasing procedural requirements in order to enhance image capturing conditions.


\section{Related works}
\label{revisao literatura}

Counting larvae of aquatic animals is a demanding, time consuming task. In aquaculture, many works have been done on counting using deep learning, both regarding adult fish and their larvae, and also other aquatic animals. For instance, \citet{hong2022underwater} proposed the use of a region-based convolutional neural network (R-CNN) with a ResNet101 backbone to identify and count fish in underwater photos. The experiment achieved an accuracy of 95.30\% on the validation dataset, despite facing some difficulties in segmenting regions of interest and in separating fish that were spatially close in the image. \citet{yu2022counting} proposed a deep learning model for fish counting called MAN, which was developed based on multiple modules and an attention mechanism. The authors achieved a counting precision of 97.12\% in different aquaculture environments. Finally, \citet{li2023lightweight} proposed a lightweight neural network for fry fish counting, reporting an MAE of 3.33.

Some works have been done on larvae counting of different aquatic animals. \citet{kakehi2021identification} proposed the use of deep learning to identify oyster larvae, with an ultimate goal of counting them, reporting an f-measure of 86.4\%. \citet{hu2023a_deep_learning_based} proposed the use of deep neural networks for density estimation to count white shrimp larvae, reporting a MAPE of 2.18\%. \citet{liu2023shrimpseed_net} proposed Shrimpseed\_Net, a neural network for counting shrimp seed, reporting a MAE of 17.28. \citet{rothschild2023computer} proposed the use of an adapted YOLOv5 model for crustacean larvae counting in different phases.

In aquaculture, the interest in fish counting is not limited to fish larvae. Some of the fish counting works focus in other early stages, such as fry fish~\cite{li2023lightweight} and fingerlings~\cite{fernandes2024convolutional}, or simply on juvenile fish~\cite{krishna2023computer}. Among the works that focused on fish larvae, the one by
\citet{costa2022deep,costa2023counting} employed a set of manually annotated images containing over 6000 tilapia larvae and achieved an accuracy of 98.5\% and a mean absolute error of 1.43. In this case, this was achieved by using a complex image capturing apparatus that included a light table and Petri dishes, in spite of the proposal of a new approach.

The applicability of the proposed approaches in real-life scenarios and use cases can be considered a general concern within the field of automated fish counting. In some of the related works mentioned here, data collection involved steps aimed at enhancing image capturing conditions, in order to increase the performance of deep learning models. In this work, we evaluate four object detection neural network architectures (plus size variants) in the task of spotted sorubim and dourado larvae counting, with images captured by smartphone and less data collection requirements. We proceed by using an image tiling technique and also evaluate different tiling scales, in order to increase detection performance.


\section{Materials and methods}

\subsection{Image dataset}
\begin{figure}
    \centering
    \begin{subfigure}[]{0.49\columnwidth}
        \centering
        \includegraphics[width=\columnwidth]{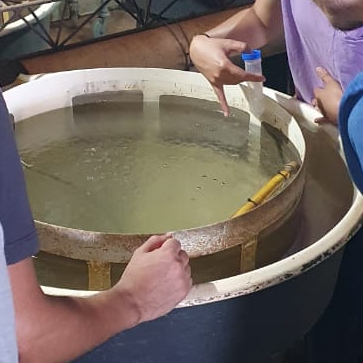}
        \caption{}
        \label{fig:coleta_pipeta}
    \end{subfigure}
    \hfill
    \begin{subfigure}[]{0.49\columnwidth}
        \centering
        \includegraphics[width=\columnwidth]{bacias\_recortada.png}
        \caption{}
        \label{fig:coleta_bacias}
    \end{subfigure}
    \hfill
    \caption{An illustration of the sampling process used to obtain the larva bowls that were photographed, along with some of the bowls available for this purpose and utilized in this study.}
    \label{fig:coleta}
\end{figure}

For this work, a new image dataset was created. It contains 162 images of small trays with spotted sorubim (\textit{Pseudoplatystoma corruscans}), dourado (\textit{Salminus brasiliensis}) and streaked prochilod (\textit{Prochilodus lineatus}) larvae. Images were collected in Projeto Pacu, an aquaculture company that specializes in large scale fish reproduction located in Terenos, Mato Grosso do Sul, Brazil. Data collection was performed by sampling, according to the following procedure. In virtue of the business process, all larvae are stored in hatcheries, from which a sample is collected using a 50 ml falcon tube. The sample is then transferred to one of a variety of common bowls dedicated to this purpose. Figure~\ref{fig:coleta_pipeta} illustrates the process of sampling with a tube and Figure~\ref{fig:coleta_bacias} shows some of the bowls used in this study.

\begin{figure}[!htb]
	\centering
    \includegraphics[width=0.7\textwidth]{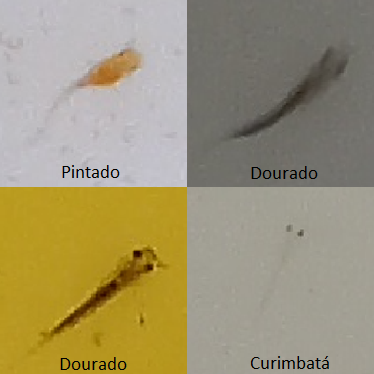}
    \caption{Close-up examples of the larvae used in this study. Streaked prochilods were not counting targets in this study, and it is not intuitively obvious that the same techniques used here will have the same efficacy in counting them as they have in counting spotted sorubim and dourado larvae. Nonetheless, it should be clear from this example that they can be treated as noise.}
    \label{fig:especies.png}
\end{figure}

For image acquisition, a short video of the bowl is taken after the sampling process. Then, one frame per video, and one frame per sample therefore, is selected based on its visual quality. The main criteria for selection is ocurrence and quantity of blurred sections. After being selected, the images were annotated with bounding boxes for object detection with only one generic class, fish, for spotted sorubim and dourado larvae. In this work, we focus on these two species. Larvae of streak prochilod were not annotated. This option was due mainly to the fact that the visual aspect of these larvae differs from that of the other ones, in that it tends to be transparent. A close-up example can be seen in Figure~\ref{fig:especies.png}, along with examples of other species. Given that this is so, it is reasonable to assume that it will require a more specific procedure. Furthermore, this species is used as fish food in dourado production, and while counting them may be useful, it is not the focus of this study. Therefore, in this work streaked prochilod larvae were considered noise, and shall be dealt with in future works.

\begin{figure*}
    \centering
    \begin{subfigure}[]{0.24\columnwidth}
        \centering
        \includegraphics[angle=90,width=\columnwidth]{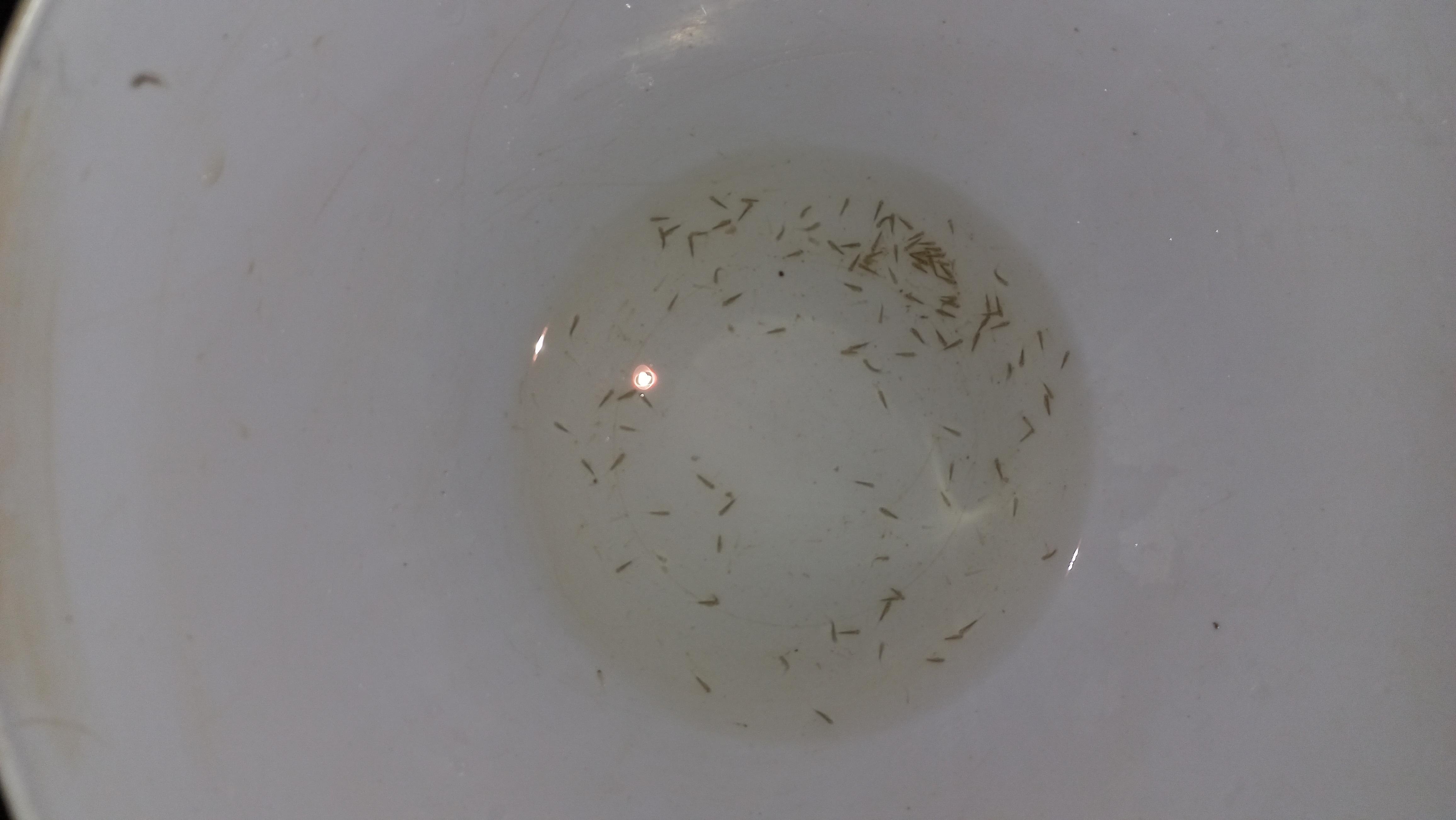}
        \caption{}
    \end{subfigure}
    \begin{subfigure}[]{0.24\columnwidth}
        \centering
        \includegraphics[width=\columnwidth]{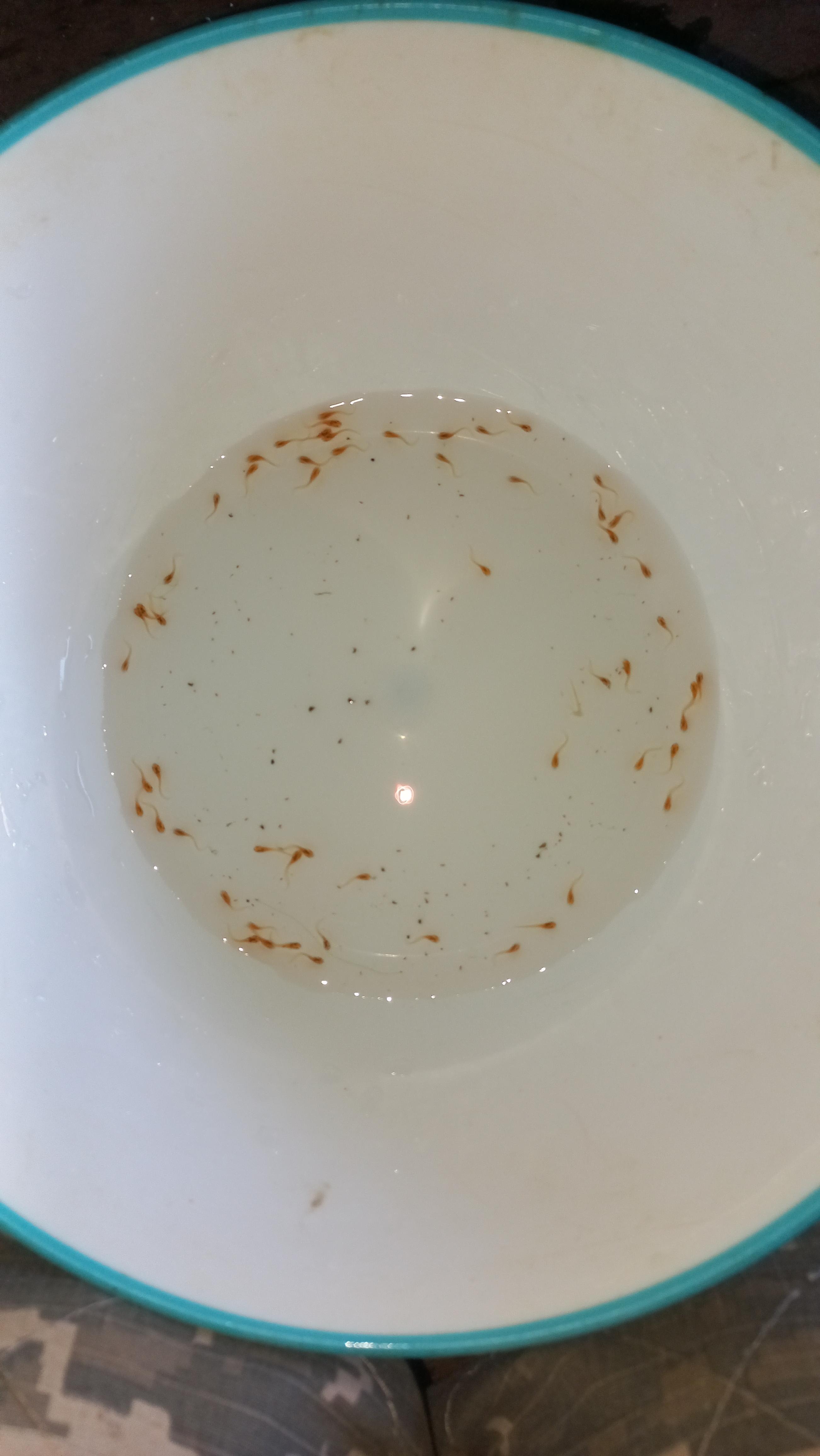}
        \caption{}
    \end{subfigure}
    \begin{subfigure}[]{0.24\columnwidth}
        \centering
        \includegraphics[width=\columnwidth]{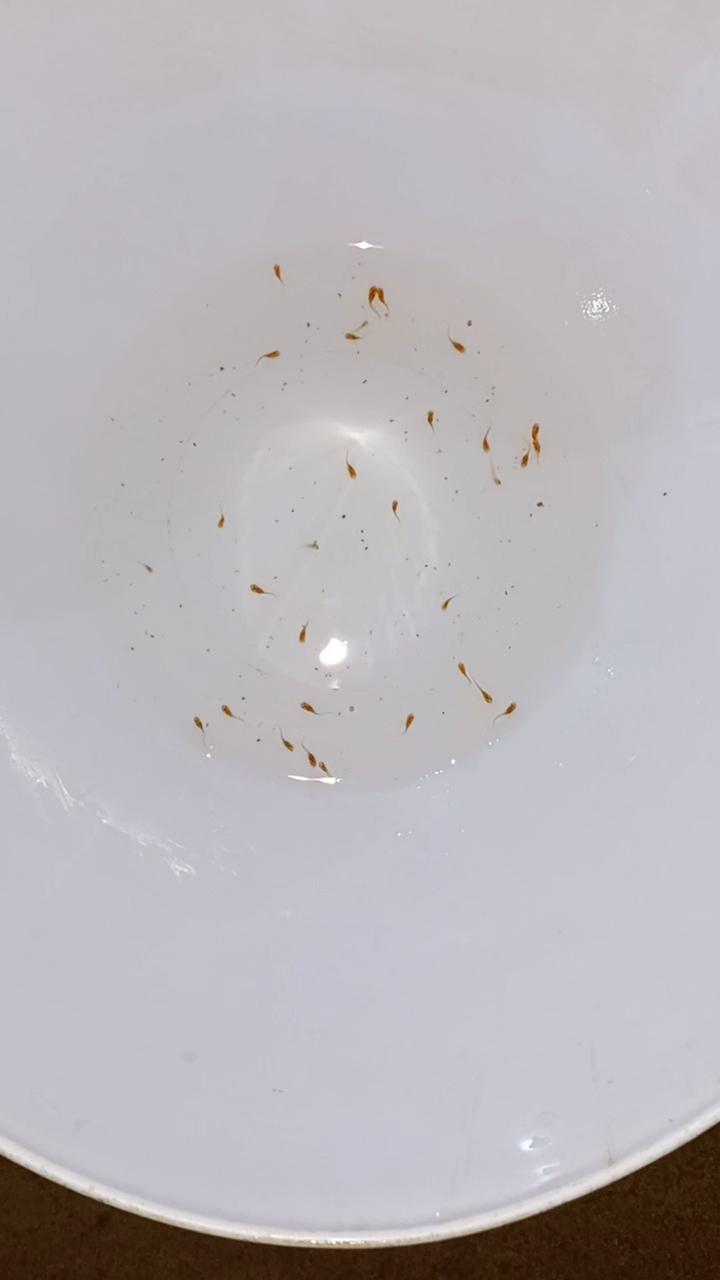}
        \caption{}
    \end{subfigure}
    \begin{subfigure}[]{0.24\columnwidth}
        \centering
        \includegraphics[width=\columnwidth]{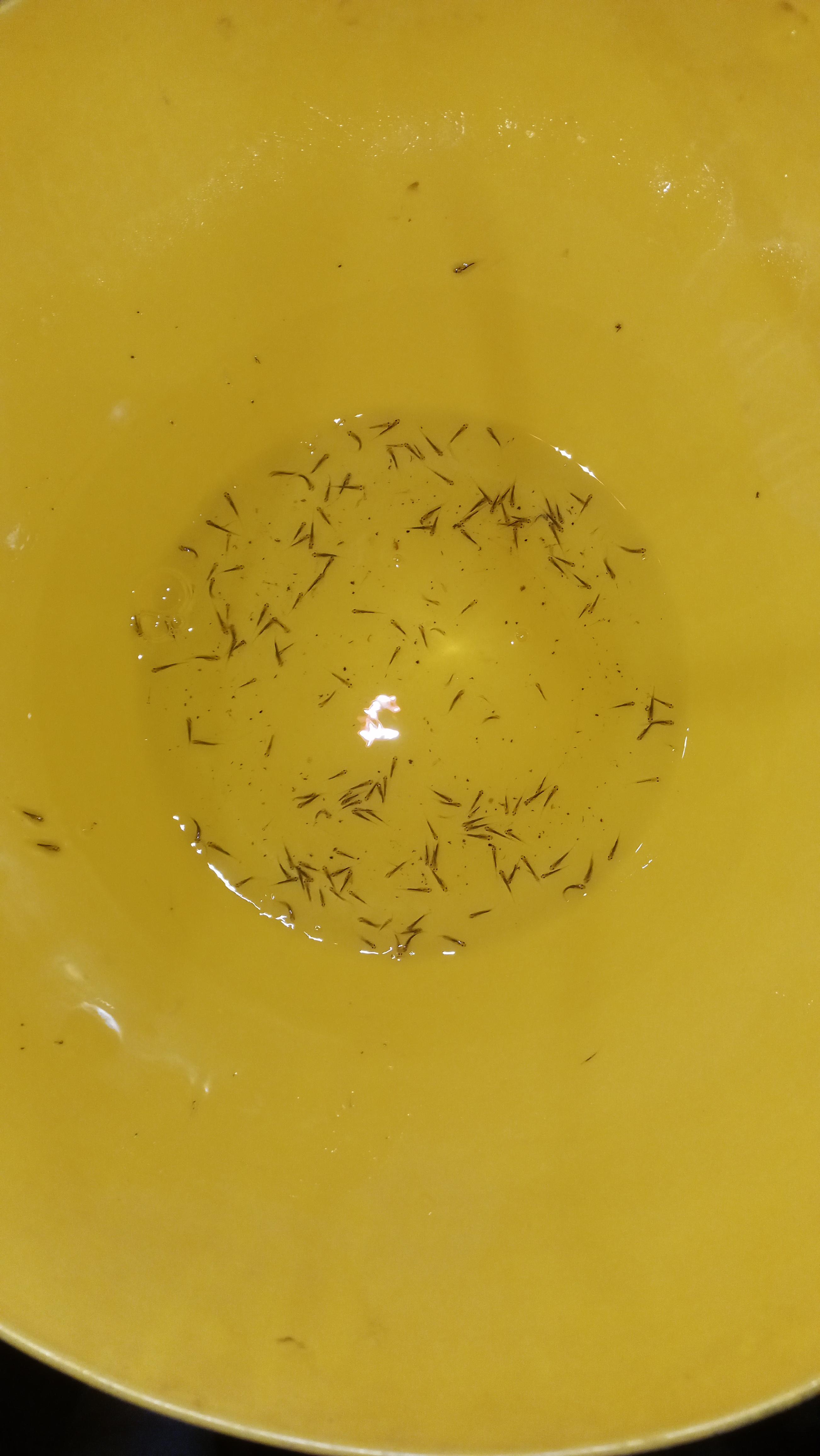}
        \caption{}
    \end{subfigure}
    \caption{Some of the images that constitute the full dataset. One should notice that, since there was no control for obtaining images in ideal and standardized conditions, the images in the dataset present some degree of variability in lighting conditions and background, as well as dirt and the dark eyes of streaked prochilod larvae.}
    \label{fig:exemplos_imagens}
\end{figure*}

The resulting images have resolution of (2604, 4624). They were captured by a specialized aquaculture professional using a Samsung SM-A325M smartphone. As described, the data collection process was not made under idealized laboratory conditions, except for the choice of plain bowls and the sampling methodology itself. This leads to the presence of noise in the images, such as dirt, background differences, and the presence of different species in the same sample. Figure~\ref{fig:exemplos_imagens} shows some of the images collected in this study.

\subsection{Image tiling} 
\label{tiling}
\label{tiling de imagens}

An image tiling strategy was used for processing with the neural networks, mainly due to hardware limitations in two senses. First, regarding the hardware available for this study. Second, and most importantly, regarding limited hardware, such as smartphones, that can be used in real-life scenarios. In this work, two strategies were used for image tiling. The first one consisted in extracting fixed-size fragments of shape (640, 640), and was used in the training phase of the neural networks.

\begin{figure}[htb!]
	\centering
    \includegraphics[width=0.7\textwidth]{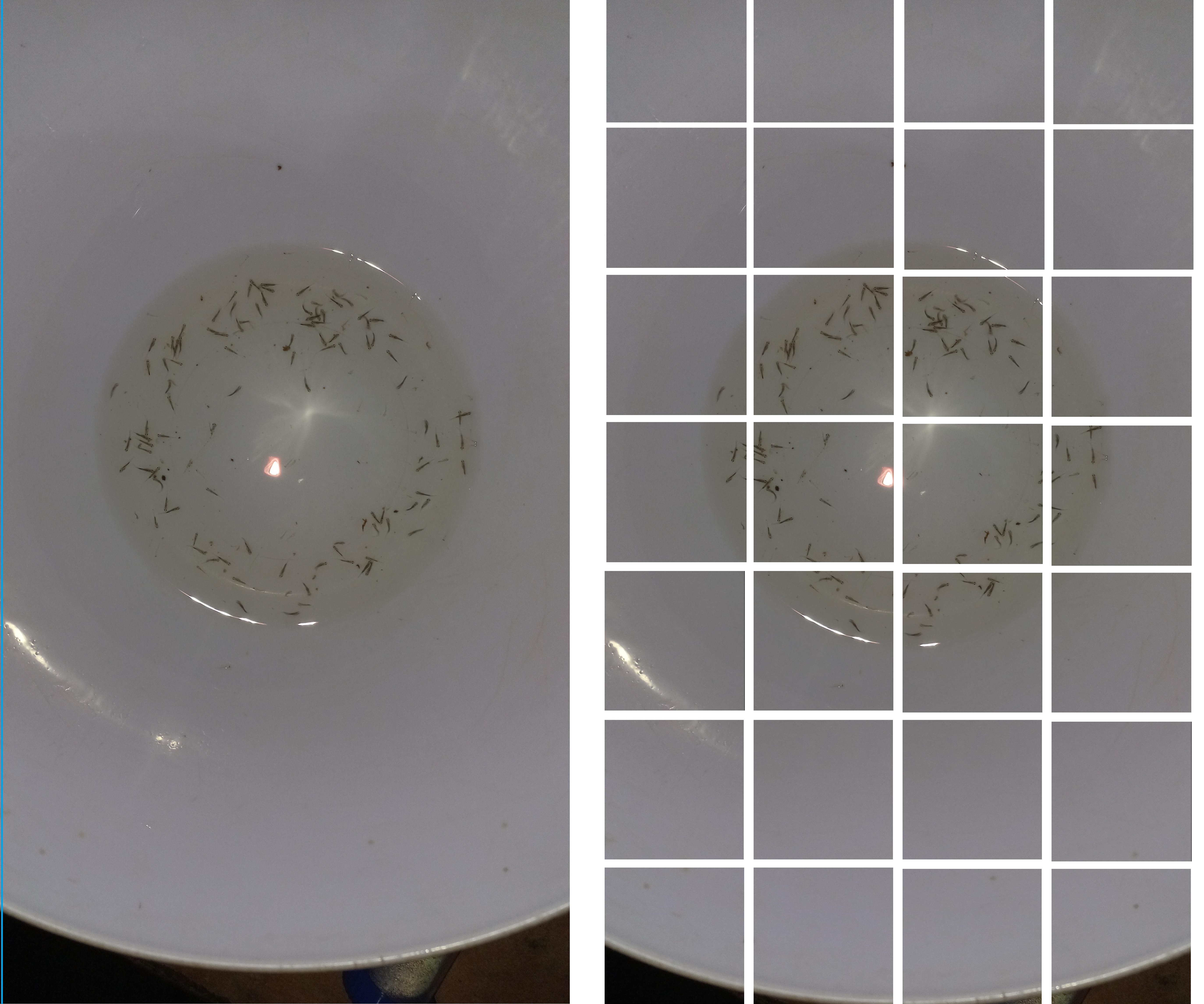}
    \caption{An example of fixed-size tiling applied on an image of the dataset. Another strategy used in this work was tiling according to a scale factor.}
    \label{fig:exemplo_tiling}
\end{figure}

The second strategy is to use a scaling factor, which gives the size of the fragments as \(ceil(min(H, W)*(s))\), where $H$ and $W$ are the height and the width of the image, respectively, and $s$ is the scaling factor. That is, each fragment is a square whose sides are given by the ceil function of the product between the smallest side of the image and the scaling factor. The image is padded with black pixels, if necessary. This second strategy was used in a second training step, in order to adjust the scaling factor, and was also used during the testing phase. Figure~\ref{fig:exemplo_tiling} shows an example of fixed-size tiling in an image of the dataset.

Since the tiled images were annotated for object detection, and tiling did not consider whether or not an annotation would span two fragments, a rule was established, according to which an annotation that went beyond the boundaries of the fragment would only be kept if at least 60\% of its area was contained within the fragment.

\subsection{Deep learning}

In this work, we evaluate four neural network architectures in the task of fish larvae counting, including convolution- and transformer-based architectures. We also test size variations, summing up to nine models evaluated. Next, we comment briefly on each architecture, and present the different versions utilized.

The first architecture evaluated by us is YOLOv8, a state-of-the-art convolutional neural network, focused on accuracy and real-time detection~\cite{jocher2023yolo}. One of its differentials is that it doesn't use the anchoring method, which is accomplished by directly predicting the center of the object instead of the displacement from a known anchor box. This simplifies the process and speeds up Non-Maximum Suppression (NMS), a complex post-processing step.

The second architecture we evaluated in this work is the detection transformer (DETR) \cite{carion2020end}, an object detection network based both on transformers~\cite{vaswani2017attention,dosovitskiy2020image} and convolutions. DETR initially uses ResNet, a CNN, as its backbone to extract a set of features from the input image. These features are then passed to an encoder-decoder Transformer, which uses attention mechanisms to capture the global relationships between the features throughout the image.

As a third architecture, we tested the Deformable DETR, an evolution of DETR that introduced a deformable attention mechanism \cite{zhu2020deformable}. While DETR uses global attention, considering all positions in the image uniformly, Deformable DETR allows the model to focus dynamically on a set of positions of interest. This is achieved through deformable attention modules that adapt the model's focus area during training. With this strategy, Deformable DETR is expected to achieve higher performance than DETR, while using fewer epochs for training.

Finally, the last architecture evaluated in this work is the real-time detection transformer (RT-DETR) \cite{lv2023detrs}. The RT-DETR (Real-Time DEtection TRansformer) model is an architecture for real-time object detection. It utilizes an encoder and a query selection method that takes into account the accuracy of the overlap between predicted and actual object areas (IoU) to improve its results. It also offers the possibility to adjust the speed at which it processes images to identify objects without the need to go through the training process again.

\begin{table}[]
    \centering
    \begin{tabular}{|c|c|c|}
        \hline
        \textbf{Model} & \textbf{Variant} & \textbf{Params. (M)} \\
        \hline \hline
        \multirow{5}{*}{YOLOv8} & nano & 3.2 \\
        & small & 11 \\
        & medium & 26 \\
        & large & 44 \\
        & extra large & 68 \\
        \hline
        RT-DETR & large & 32 \\
        RT-DETR & extra large & 67 \\
        \hline
        \multicolumn{2}{|c|}{DETR-ResNet50} & 41 \\
        \hline
        \multicolumn{2}{|c|}{Deformable DETR} & 40 \\
        \hline
    \end{tabular}
    \caption{Number of parameters (in Millions) for each architecture evaluated in this work, specified by variant (when applicable).}
    \label{tab:num_params}
\end{table}

\subsection{Experimental design}
\label{exp_des}

In order to evaluate the Neural Networks for object detection in the proposed approach, the dataset was divided into training (60\%), validation (20\%) and test (20\%) sets, and the experiment followed a three-step procedure. First, the object detection models were trained and validated for fish larvae detection. In this step, all models were fine-tuned without freezing any layers. For training, the maximum number of epochs was set to 100 for the Deformable DETR, and 400 for the RT-DETR and also for the YoloV8 models. Furthermore, early stopping was used with 50 epochs of patience. Regarding Deformable DETR, validation loss values were monitored. For YoloV8 and RT-DETR an average calculated on the mAP values were monitored.

After the neural networks were trained, an inference hyperparameter tuning step took place, in which the confidence thresholds and the tiling scales were combined for each neural network. Several values were iteratively tested, where each iteration was carried out on a smaller range of values, with increasing precision. This second phase was carried out on the training set. Additionally, in order to reduce the time required for computation, we downscaled the image resolution by a factor of 0.5. Since most of the images in the dataset were in vertical orientations, we also used the following data augmentation techniques: rotation between 0\textdegree and 360\textdegree without changing image orientation, clockwise rotation between 0\textdegree and 90\textdegree changing image orientation, and horizontal and vertical mirroring. In order to choose a combination of hyperparameters, we calculated the MAE, MAPE and RMSE for each combination of hyperparameters, applied min-max normalization and chose the combination that yielded the smallest sum. This same procedure is used to order the results in the following Section.

Finally, the models were tested with the hyperparameters chosen for each model in the preceding step. In order to evaluate the results, we used the Analysis of Variance (ANOVA), followed by TukeyHSD, by considering each image a repetition, both at 5\% significance threshold. Given this procedure, the hypothesis test was applied on the absolute difference and on the absolute percentage difference (with which MAE and MAPE are calculated, respectively). We also calculated the coefficient of determination ($R^2$), and generated scatterplots.

\section{Results and discussion} 
\label{resultados}

Table~\ref{tabela-conf-tiling} shows the confidence thresholds and the tiling scales chosen in the post-training hyperparameter tuning step for each model evaluated in this work. It is possible to see that both the confidence threshold and the tiling scale varied considerably, ranging from 25\% up to 70\%, and from 20\% up to 50\%, respectively.

\begin{table}[htb!]
    \centering
    \begin{tabular}{|c|c|c|} \hline  
        \textbf{Model} & \textbf{Conf. threshold} & \textbf{Tiling scale} \\ \hline \hline
        yolov8n & 45\%& 40\%\\ \hline  
        yolov8s &     40\%& 50\%\\ \hline  
        yolov8m &     40\%&     50\%\\ \hline  
        yolov8l & 50\%&30\%\\ \hline  
        yolov8x &     45\%&     30\%\\ \hline  
        rtdetr-l & 70\%&20\%\\ \hline  
        rtdetr-x & 50\%&50\%\\ \hline  
        detr-resnet-50& 40\%&30\%\\ \hline  
        deformable-detr& 25\%&20\%\\ \hline 
    \end{tabular}
    \caption{Confidence thresholds and tiling scales chosen for each model in the hyperparameter tuning step.}
    \label{tabela-conf-tiling}
\end{table}

Table~\ref{resultados_finais} shows average MAE, MAPE and RMSE results in the test set, along with their standard deviations, for each architecture evaluated, ordered by smallest sum of normalized means. Table~\ref{resultados_finais} also shows the results of the TukeyHSD in compact letter display format. One can see that rtdetr-x achieved the smallest average MAE and MAPE on the test set, with the smallest standard deviation on the later, and second smallest on the former. However, yolov8m achieved the smallest average RMSE, and also the smallest sum of normalized means. The ANOVA test was indicative of statistically significant difference for both MAE and MAPE, with $p = 0.00144$ for MAE, and $p = 0.00164$ for MAPE. On the other hand, the post-hoc TukeyHSD applied did not indicate that either yolov8m or rtdetr-x differed from the other models, exception made to the detr-resnet-50.

\begin{table*}[htb!]
    \centering
    \begin{tabular}{|c|c|c|c|c|}
        \hline
        Model & MAE & MAPE & RMSE & $R^2$\\ \hline \hline
        yolov8m & 5.44±6.81 a & 4.71±4.98 a & \textbf{8.63±11.96} & \textbf{0.983} \\ \hline
        rtdetr-x & \textbf{5.41±7.52 a} & \textbf{4.46±4.70 a} & 9.17±13.87 & 0.981 \\ \hline
        yolov8n & 7.31±10.14 a & 6.00±6.53 a & 12.37±18.73 & 0.966 \\ \hline
        yolov8x & 7.81±9.95 a & 6.66±6.96 a  & 12.53±17.87 & 0.965 \\ \hline
        rtdetr-l & 8.31±9.33 ab & 7.45±6.14 ab & 12.39±18.12 & 0.966 \\ \hline
        yolov8s & 8.44±12.12 ab & 7.19±8.64 ab & 14.61±21.91 & 0.953 \\ \hline
        yolov8l & 9.72±12.06 ab & 7.67±8.26 ab & 15.34±21.49 & 0.948 \\ \hline
        deformable-detr & 14.47±17.10 ab & 13.85±12.48 ab & 22.19±32.44 & 0.891 \\ \hline
        detr-resnet-50 & 18.47±25.79 b & 18.74±38.86 b & 31.39±47.74 & 0.781 \\ \hline
        \end{tabular}
    \caption{Average MAE, MAPE, RMSE and $R^2$ results in the test set, along with their respective standard deviations, for each neural network evaluated in this work. The results are ordered smallest sum of normalized means, according to the procedure described in Section~\ref{exp_des}. TukeyHSD results can be seen in compact letter display format within MAE and MAPE columns.}
    \label{resultados_finais}
\end{table*}

\begin{figure}[htb!]
	\centering
    \includegraphics[width=\columnwidth]{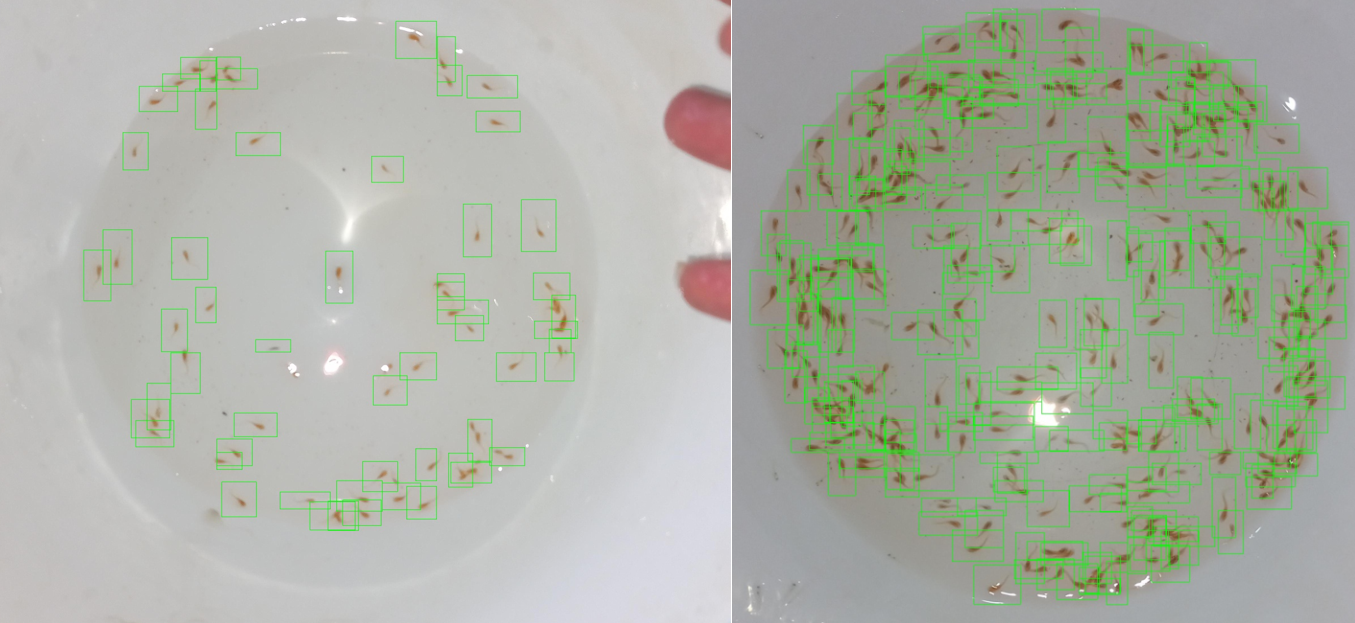}
    \caption{Examples of larvae counting with YOLOv8n. In these images, the percentage errors achieved were 5\% and 2.41\%.}
    \label{fig:resultados_contagem.png}
\end{figure}

\begin{figure*}
     \centering
     \begin{subfigure}[b]{0.3\textwidth}
         \centering
         \includegraphics[width=\textwidth]{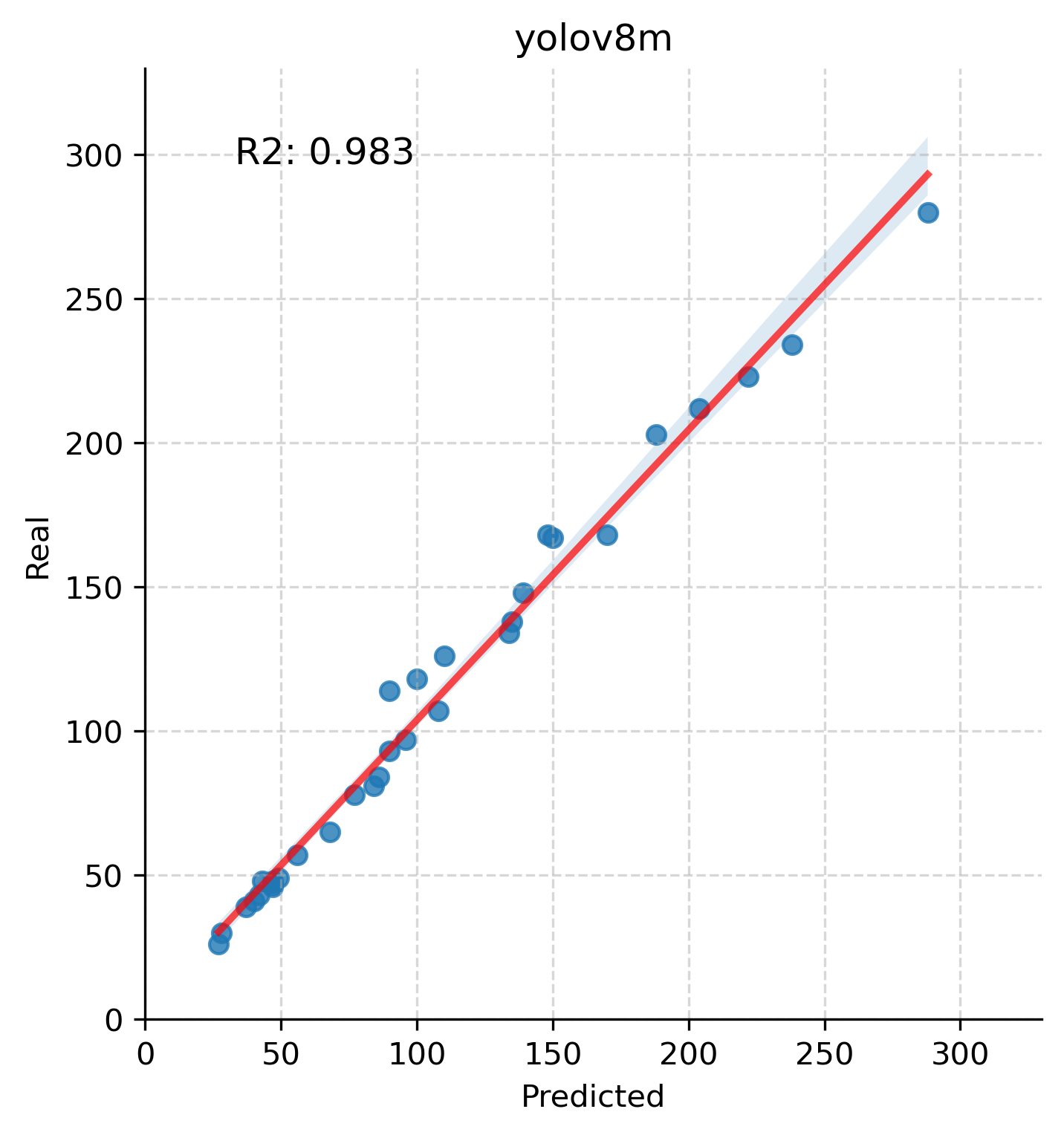}
     \end{subfigure}
     \hfill
     \begin{subfigure}[b]{0.3\textwidth}
         \centering
         \includegraphics[width=\textwidth]{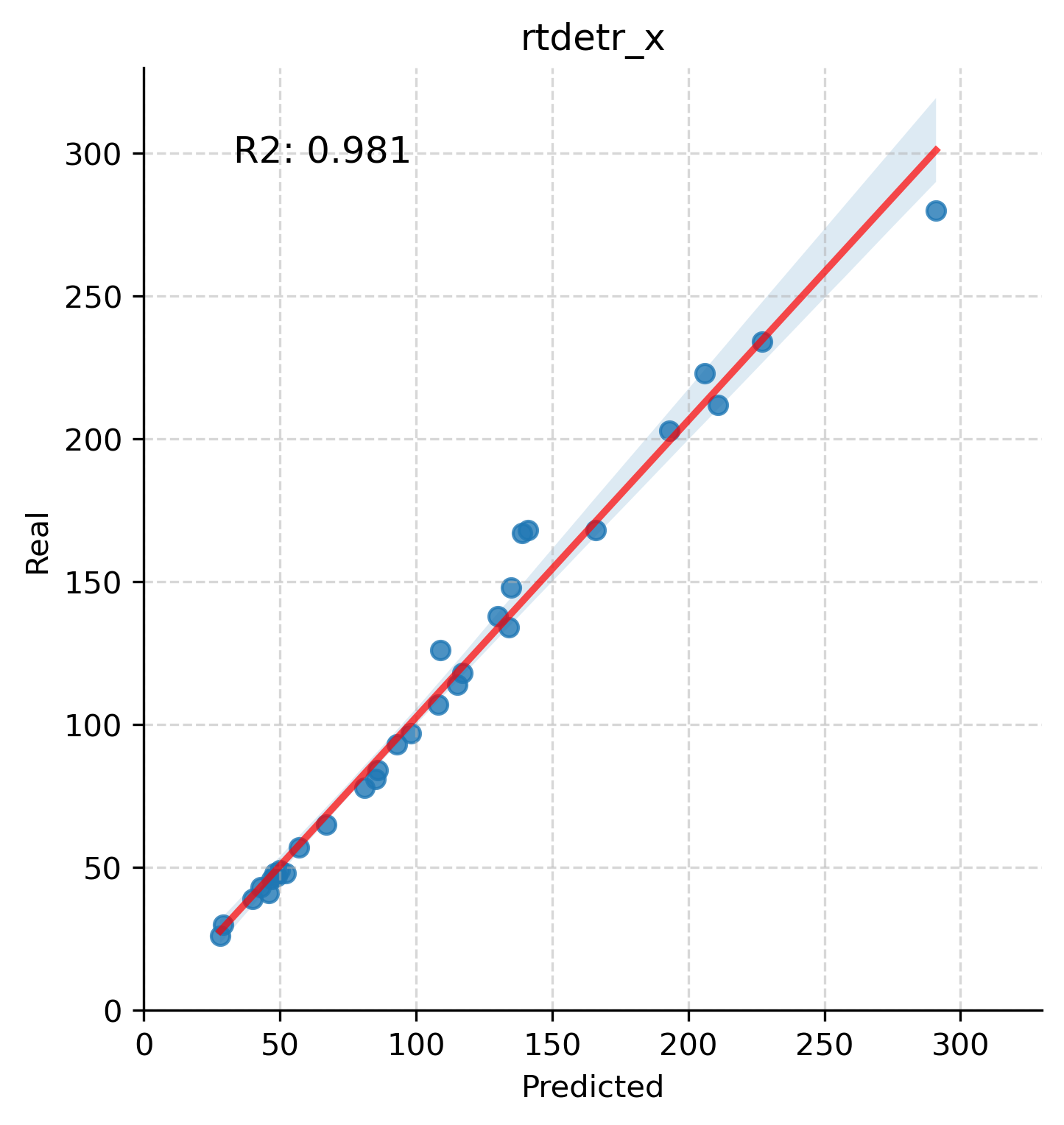}
     \end{subfigure}
     \hfill
     \begin{subfigure}[b]{0.3\textwidth}
         \centering
         \includegraphics[width=\textwidth]{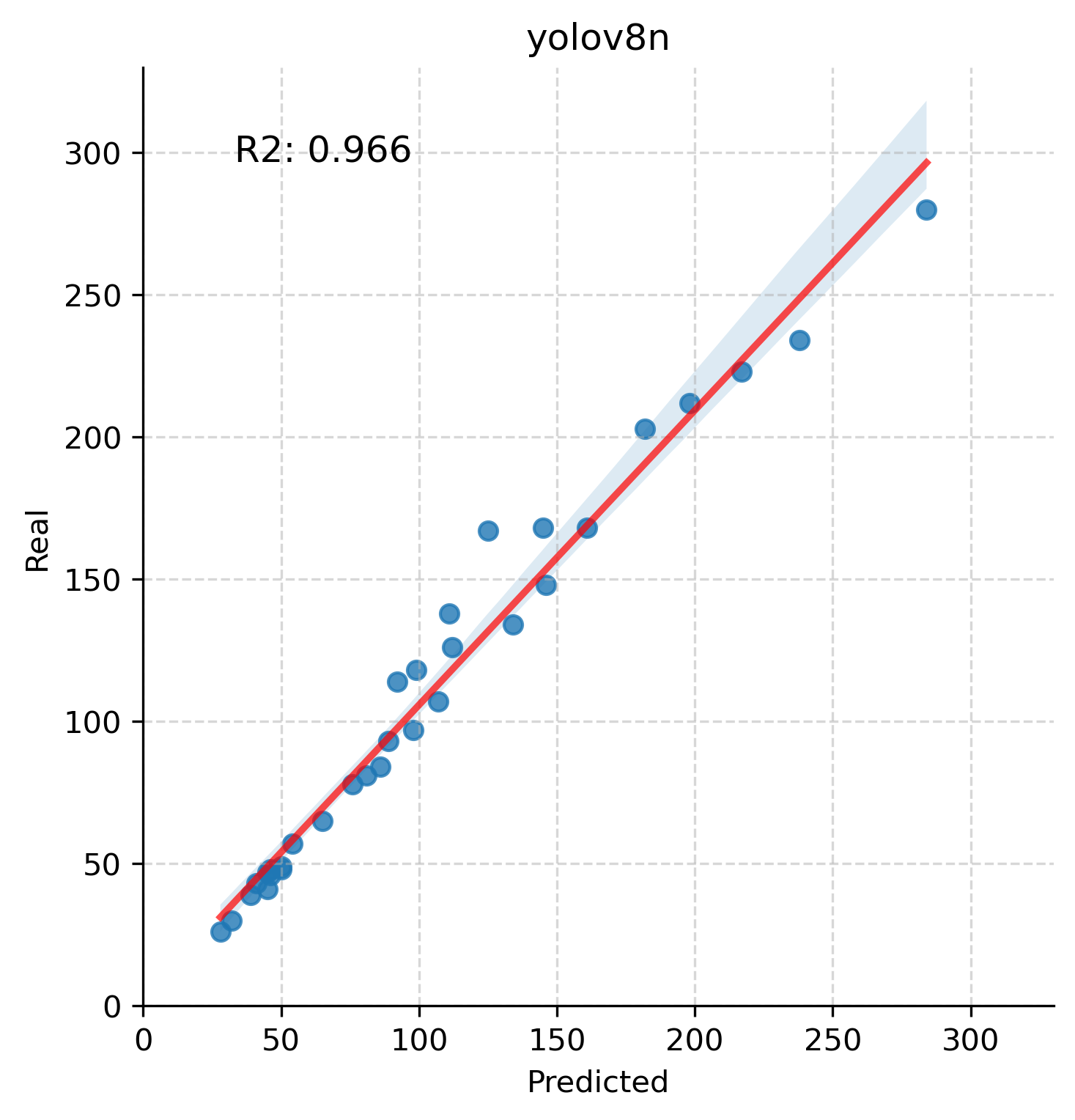}
     \end{subfigure}
     \hfill
     \begin{subfigure}[b]{0.3\textwidth}
         \centering
         \includegraphics[width=\textwidth]{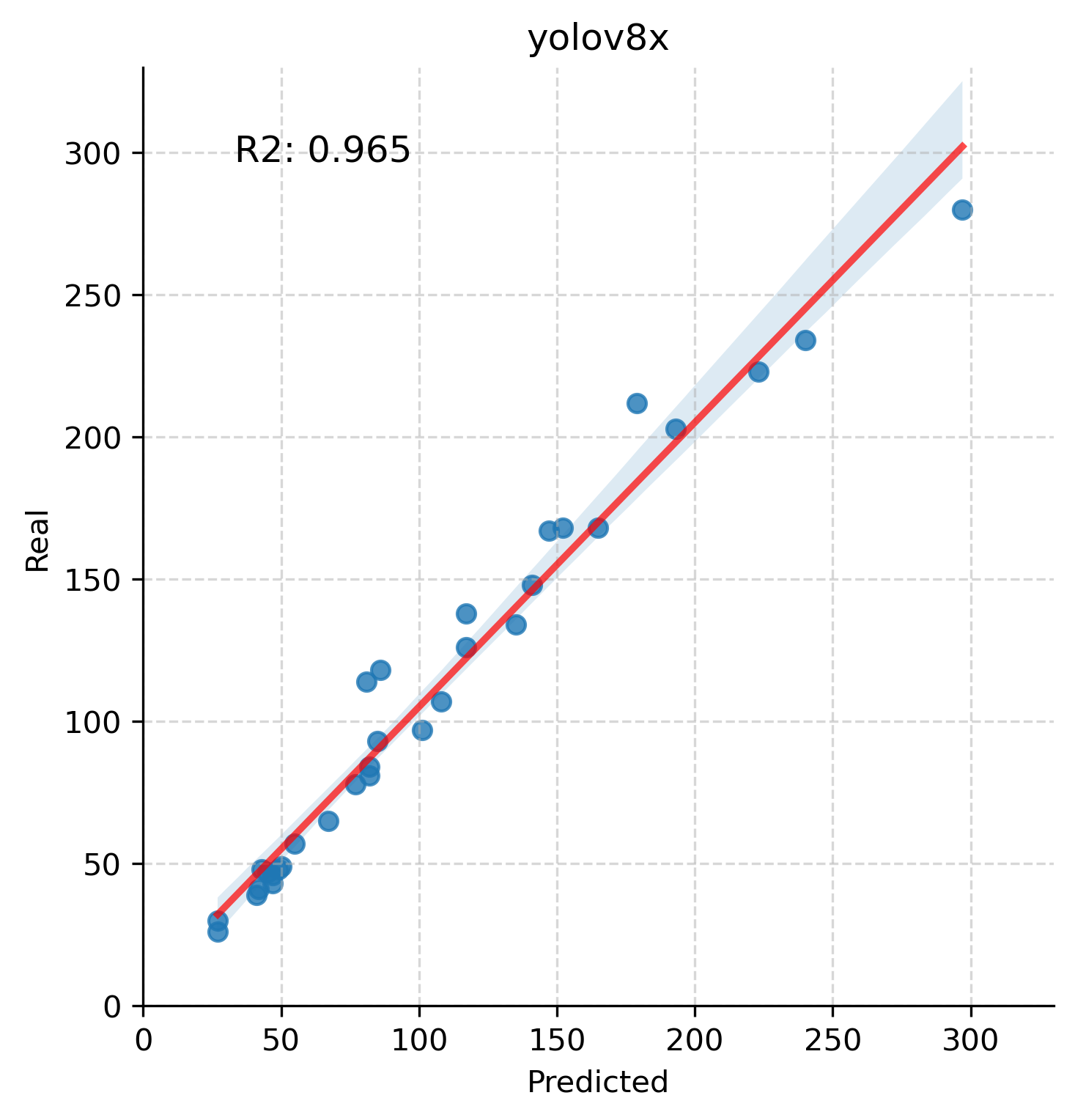}
     \end{subfigure}
     \hfill
     \begin{subfigure}[b]{0.3\textwidth}
         \centering
         \includegraphics[width=\textwidth]{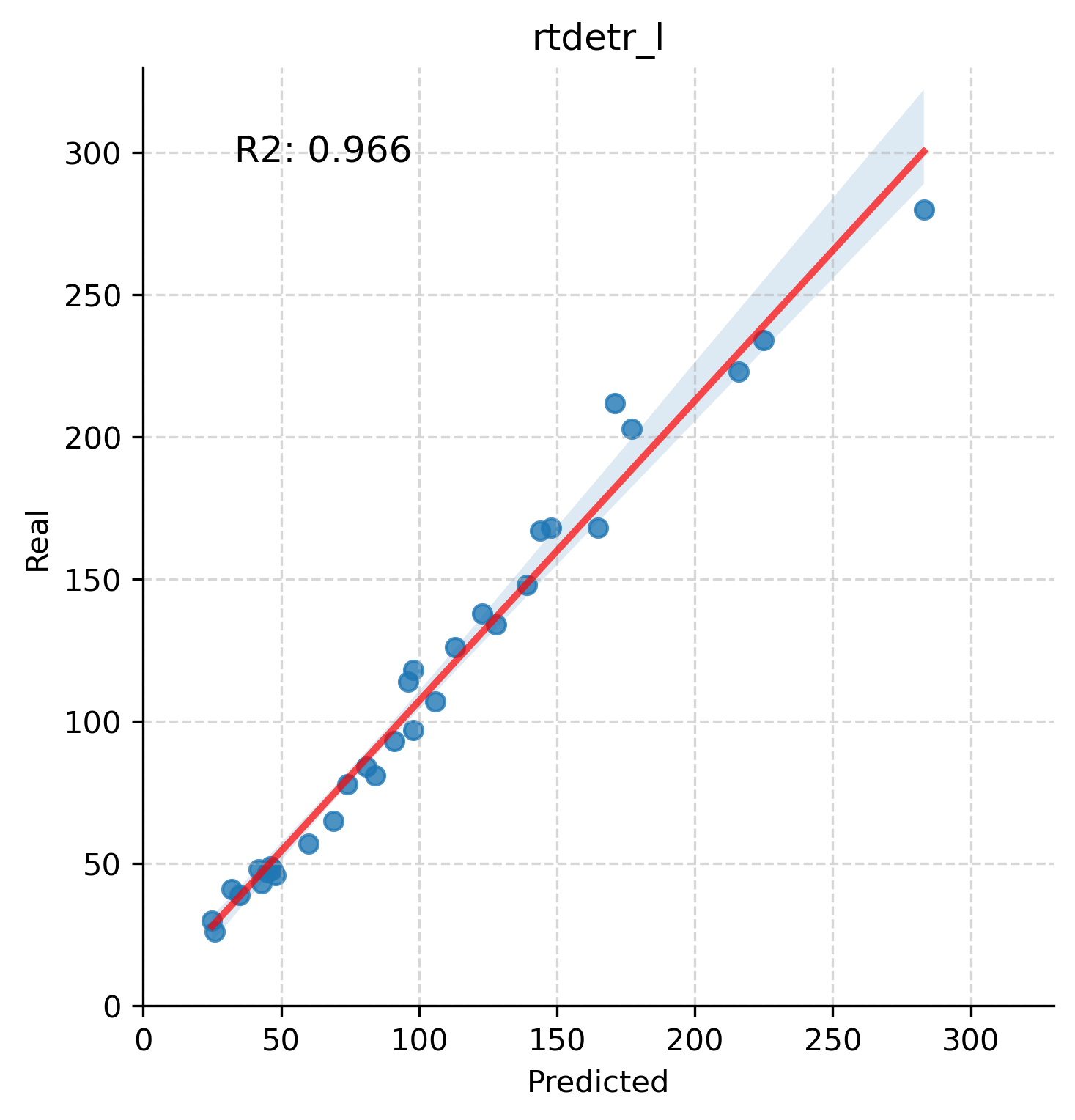}
     \end{subfigure}
     \hfill
     \begin{subfigure}[b]{0.3\textwidth}
         \centering
         \includegraphics[width=\textwidth]{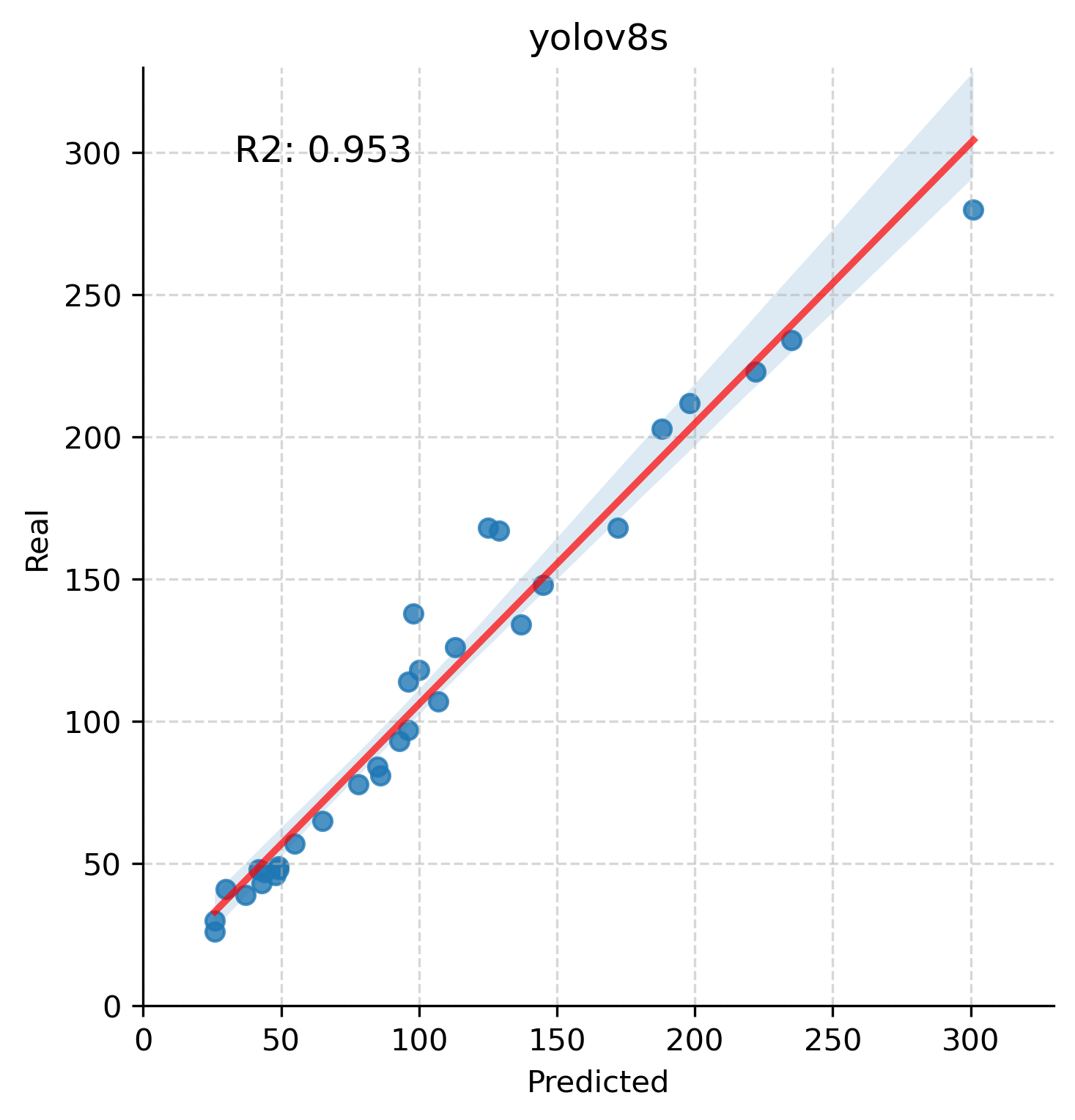}
     \end{subfigure}
     \hfill
     \begin{subfigure}[b]{0.3\textwidth}
         \centering
         \includegraphics[width=\textwidth]{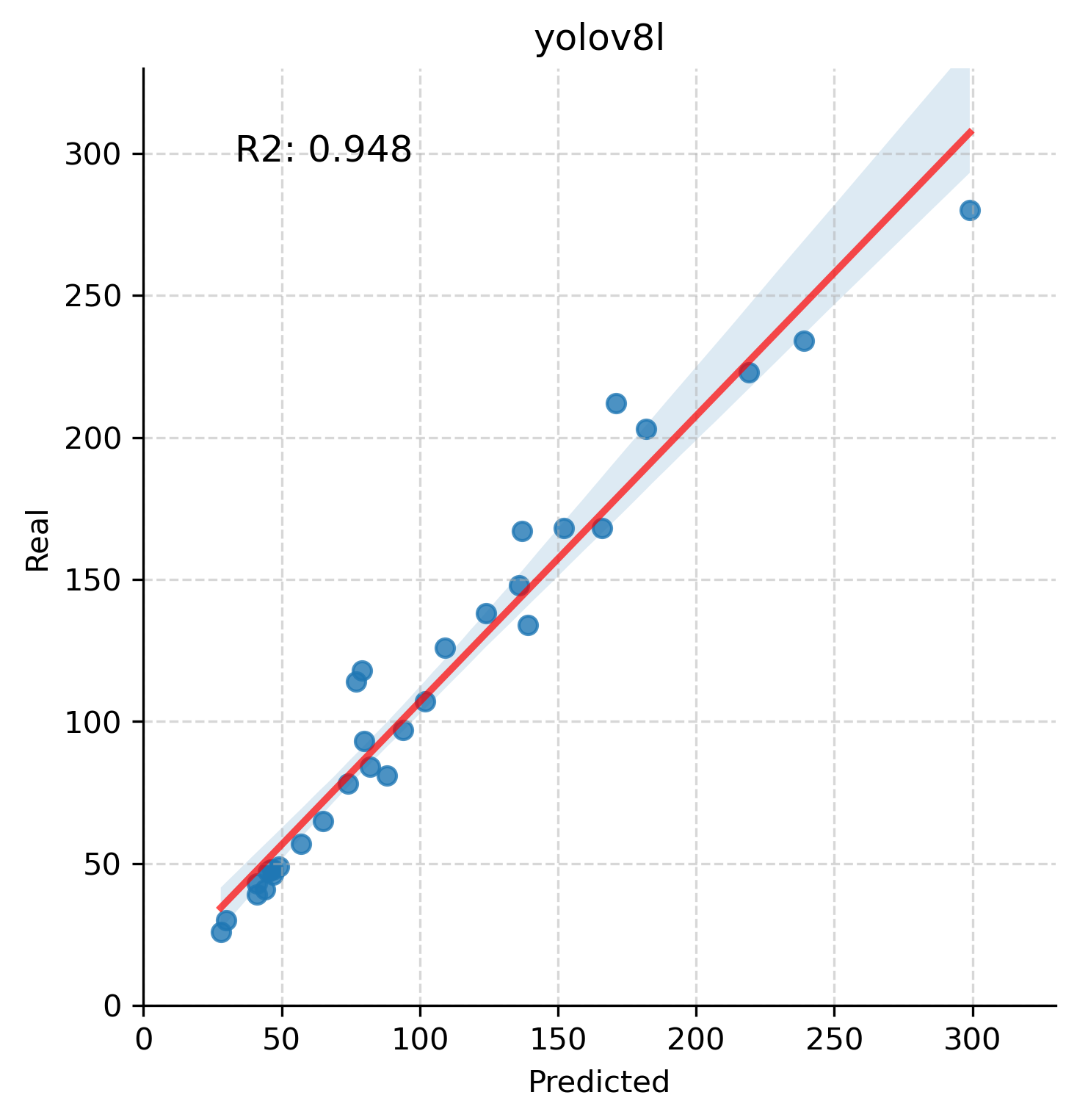}
     \end{subfigure}
     \hfill
     \begin{subfigure}[b]{0.3\textwidth}
         \centering
         \includegraphics[width=\textwidth]{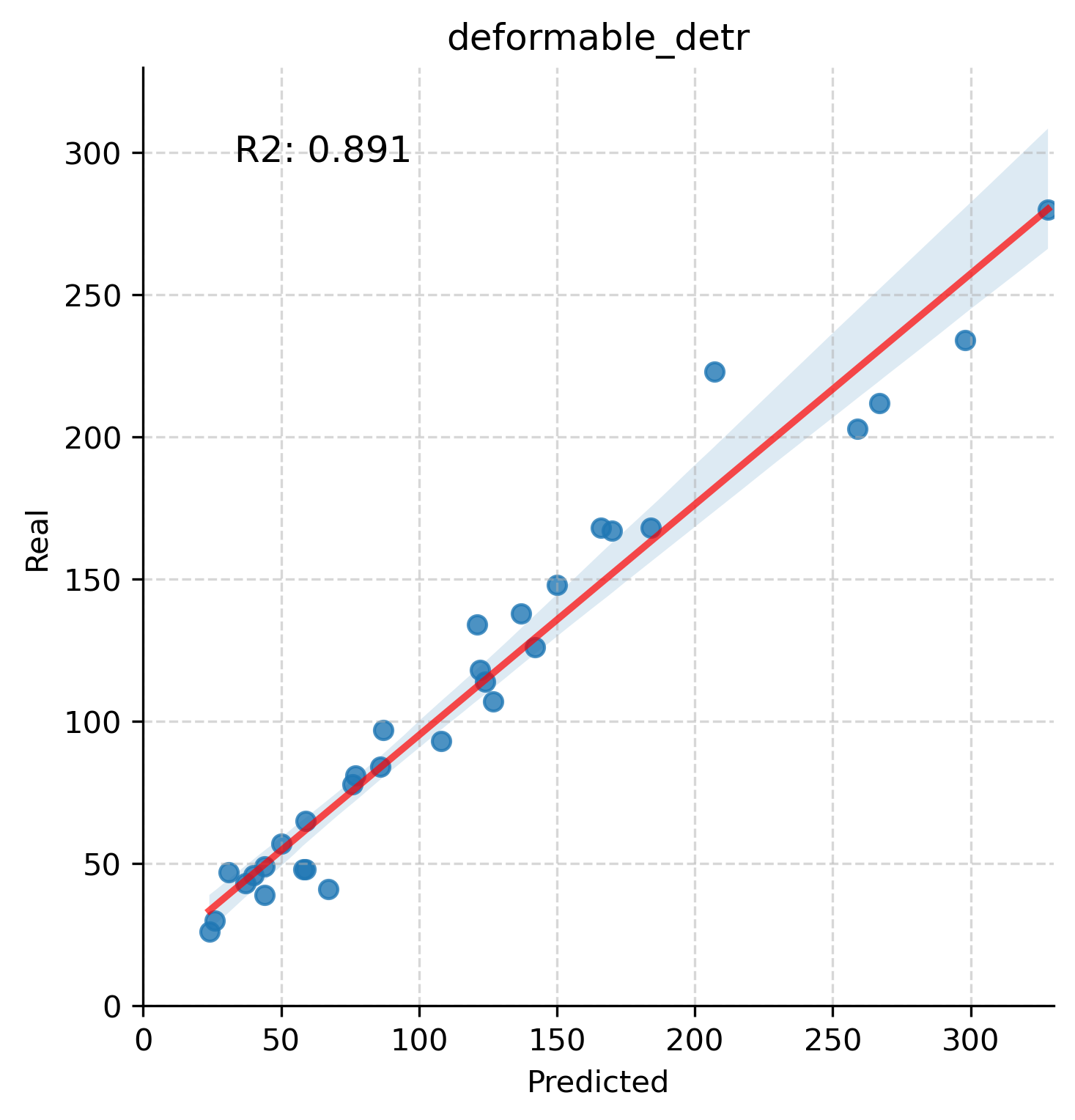}
     \end{subfigure}
     \hfill
     \begin{subfigure}[b]{0.3\textwidth}
         \centering
         \includegraphics[width=\textwidth]{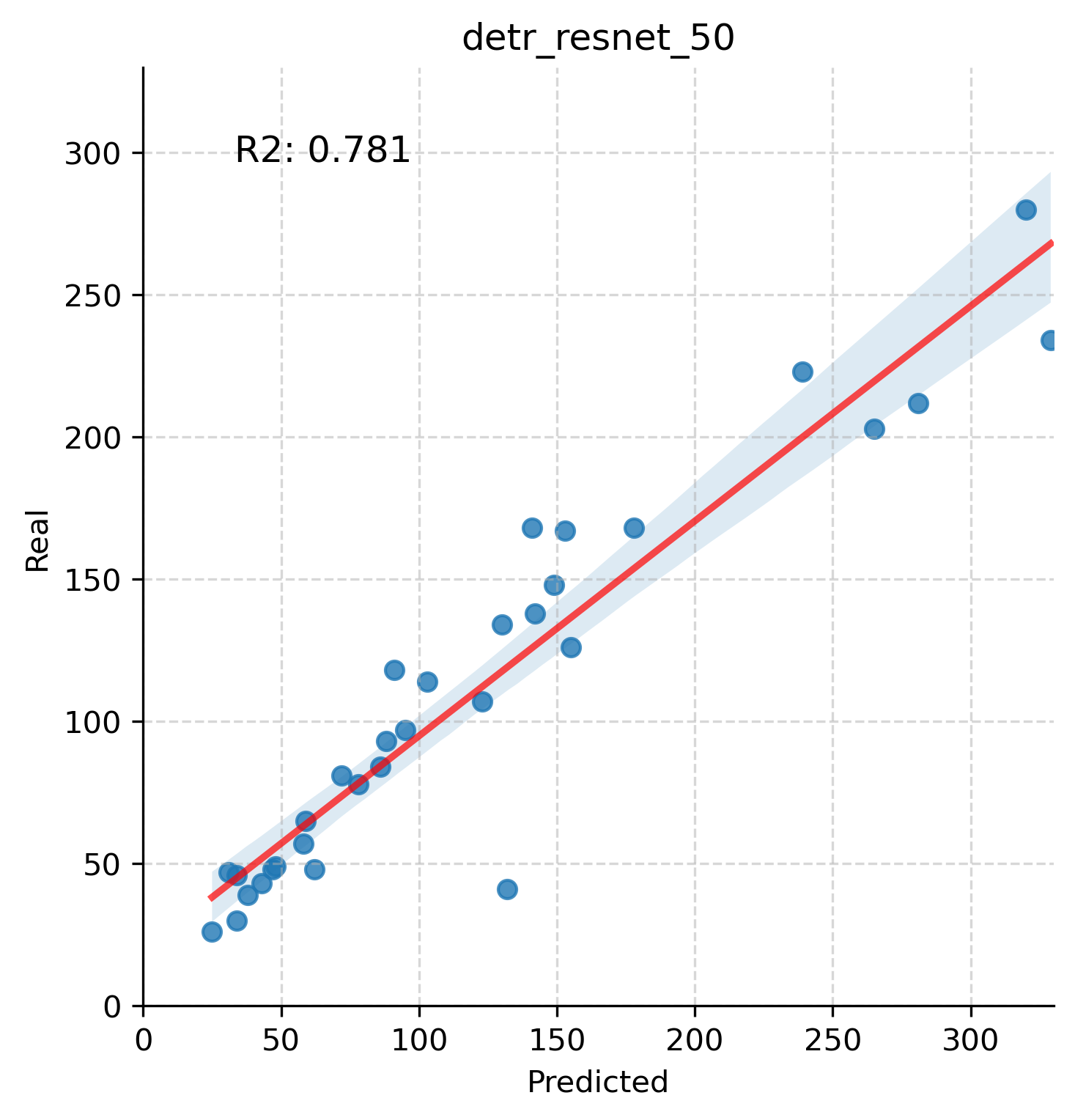}
     \end{subfigure}
        \caption{Scatterplots of groundtruth counting and predictions in the test set for each model evaluated in this study, along with their coefficients of determination.}
        \label{fig:scatterplots}
\end{figure*}

Figure~\ref{fig:scatterplots} shows the scatterplots of test predictions, compared to the groundtruth counting, along with the coefficient of determination. These plots, along with Table~\ref{resultados_finais}, show that most models achieved an $R^2$ close to 1, which indicates that most models achieved a strong linear correlation. The nano variant of the YOLOv8 architectures achieved an $R^2$ higher than larger variants, albeit smaller than the $R^2$ achieved by the medium version. As indicated by the error measurements, the Deformable DETR and the DETR with ResNet50 backbone showed a smaller $R^2$ value, which is indicative of an inferior adjustment. Finally, the medium version of YOLOv8 and the extra large version of RT-DETR showed the highest $R^2$ values. Visually, one can see that they also presented the smallest deviation from the best fit line, especially when higher counting numbers are considered.

\begin{figure}[htb!]
	\centering
    \includegraphics[width=0.8\textwidth]{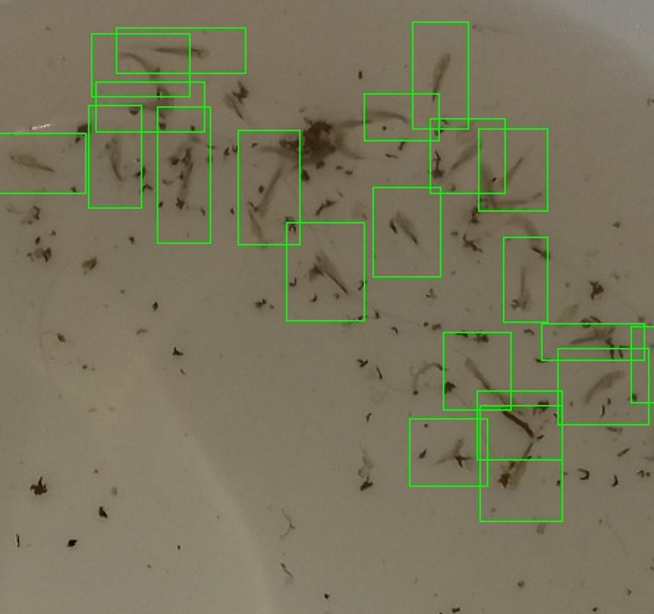}
    \caption{Larvae detection by YOLOv8n with presence of dirt and other elements that can be misclassified as larvae, in virtue of their color and shape. In this example, the percentage error of the model was 17.31\%, which can be considered high, when compared with the MAPE found in this work (see Table~\ref{resultados_finais}).}
    \label{fig:yolo_sujeiras}
\end{figure}

The results achieved were below those found by \citet{costa2023counting}. Arguably, the main reason for this is that in their case the images were taken in an idealized environment, with more clean backgrounds, higher contrast between larvae and background, better illumination, and more clean water. Their procedure, however, imposed additional requirements for image collection, and is less suitable for real-life use cases. In Figure~\ref{fig:yolo_sujeiras}, it is possible to see a close-up example that illustrates how dirt can be present in the water, in color and format that are similar to that of fish larvae, and act as noise for the neural networks. The same can be argued regarding the work by \cite{fernandes2024convolutional}, who reported recall rates of 99\% in the detection of serrasalmid fingerlings. In this case, the fish were in a more developed stage, but the image capturing was also highly standardized.

\section{Conclusion}

In this work, we performed a comparative analysis of different neural network architectures for the task of detecting and counting fish larvae. Using a set of annotated images, we explored the performance of Transformer-based models such as RT-DETR, DETR-ResNet-50 and Deformable-DETR, as well as variants of the YOLOv8 architecture. For the evaluation, we created a new annotated image dataset that imposes less data colletion requirements, whose images can be considered more representative of real-life conditions than other existing datasets for similar purposes. We achieved MAPE 4.46 ($\pm 4.70$) with an extra large RT-DETR, and 4.71 ($\pm 4.98$) with a medium sized YOLOv8. Further work can focus on increasing the dataset, as well as in evaluating different models and approaches for the task of larvae counting. In order to provide a larvae counting system for aquaculture, it may also be beneficial to develop a procedure for counting fish larvae without the need for sampling techniques.


\section{Acknowledgements}

This work has received financial support from the Dom Bosco Catholic University and from the Foundation for the Support and Development of Education, Science and Technology from the State of Mato Grosso do Sul, FUNDECT. Some of the authors have been awarded with Scholarships from the the Brazilian National Council of Technological and Scientific Development, CNPq and the Coordination for the Improvement of Higher Education Personnel, CAPES. We are especially grateful to the Projeto Pacu Ltda., for allowing and helping with data collection.






\bibliographystyle{models/elsarticle-harvard}
\clearpage


\begin{thebibliography}{20}
\expandafter\ifx\csname natexlab\endcsname\relax\def\natexlab#1{#1}\fi
\expandafter\ifx\csname url\endcsname\relax
  \def\url#1{\texttt{#1}}\fi
\expandafter\ifx\csname urlprefix\endcsname\relax\def\urlprefix{URL }\fi

\bibitem[{Barbedo(2022)}]{barbedo2022review}
Barbedo, J. G.~A., 2022. A review on the use of computer vision and artificial
  intelligence for fish recognition, monitoring, and management. Fishes 7~(6),
  335.

\bibitem[{Carion et~al.(2020)Carion, Massa, Synnaeve, Usunier, Kirillov, and
  Zagoruyko}]{carion2020end}
Carion, N., Massa, F., Synnaeve, G., Usunier, N., Kirillov, A., Zagoruyko, S.,
  2020. End-to-end object detection with transformers. In: European conference
  on computer vision. Springer.

\bibitem[{Costa et~al.(2023)Costa, Gon{\c{c}}alves, Zanoni, de~Arruda,
  de~Ara{\'u}jo~Carvalho, Nascimento, Junior, Diemer, and
  Pistori}]{costa2023counting}
Costa, C.~S., Gon{\c{c}}alves, W.~N., Zanoni, V. A.~G., de~Arruda, M. d.~S.,
  de~Ara{\'u}jo~Carvalho, M., Nascimento, E., Junior, J.~M., Diemer, O.,
  Pistori, H., 2023. Counting tilapia larvae using images captured by
  smartphones. Smart Agricultural Technology 4, 100160.

\bibitem[{Costa et~al.(2022)Costa, Zanoni, Curvo, de~Araújo~Carvalho, Boscolo,
  Signor, dos Santos~de Arruda, Nucci, Junior, Gonçalves, Diemer, and
  Pistori}]{costa2022deep}
Costa, C.~S., Zanoni, V. A.~G., Curvo, L. R.~V., de~Araújo~Carvalho, M.,
  Boscolo, W.~R., Signor, A., dos Santos~de Arruda, M., Nucci, H. H.~P.,
  Junior, J.~M., Gonçalves, W.~N., Diemer, O., Pistori, H., 2022. Deep
  learning applied in fish reproduction for counting larvae in images captured
  by smartphone. Aquacultural Engineering 97, 102225.

\bibitem[{Dosovitskiy et~al.(2020)Dosovitskiy, Beyer, Kolesnikov, Weissenborn,
  Zhai, Unterthiner, Dehghani, Minderer, Heigold, Gelly,
  et~al.}]{dosovitskiy2020image}
Dosovitskiy, A., Beyer, L., Kolesnikov, A., Weissenborn, D., Zhai, X.,
  Unterthiner, T., Dehghani, M., Minderer, M., Heigold, G., Gelly, S., et~al.,
  2020. An image is worth 16x16 words: Transformers for image recognition at
  scale. arXiv preprint arXiv:2010.11929.

\bibitem[{Fernandes et~al.(2024)Fernandes, Costa, França, Souza, Viadanna,
  Lima, Horn, Pierozan, Rezende, Medeiros, Braganholo, Silva, Nacife,
  Pinho~Costa, Silva, and Oliveira}]{fernandes2024convolutional}
Fernandes, M.~P., Costa, A.~C., França, H. F. d.~C., Souza, A.~S., Viadanna,
  P. H. d.~O., Lima, L. d.~C., Horn, L.~D., Pierozan, M.~B., Rezende, I. R.~d.,
  Medeiros, R. M. d. S.~d., Braganholo, B.~M., Silva, L. O. P.~d., Nacife,
  J.~M., Pinho~Costa, K. A.~d., Silva, M. A. P.~d., Oliveira, R. F.~d., 2024.
  Convolutional neural networks in the inspection of serrasalmids
  (characiformes) fingerlings. Animals 14~(4).

\bibitem[{Gon{\c{c}}alves et~al.(2022)Gon{\c{c}}alves, Acosta, Ramos, Osco,
  Furuya, Furuya, Li, Junior, Pistori, and
  Gon{\c{c}}alves}]{gonccalves2022using}
Gon{\c{c}}alves, D.~N., Acosta, P.~R., Ramos, A. P.~M., Osco, L.~P., Furuya, D.
  E.~G., Furuya, M. T.~G., Li, J., Junior, J.~M., Pistori, H., Gon{\c{c}}alves,
  W.~N., 2022. Using a convolutional neural network for fingerling counting: A
  multi-task learning approach. Aquaculture 557, 738334.

\bibitem[{Hong~Khai et~al.(2022)Hong~Khai, Abdullah, Hasan, and
  Tarmizi}]{hong2022underwater}
Hong~Khai, T., Abdullah, S., Hasan, M., Tarmizi, A., 2022. Underwater fish
  detection and counting using mask regional convolutional neural network.
  water 2022, 14, 222.

\bibitem[{Hu et~al.(2023)Hu, Chen, Hsieh, and
  Ting}]{hu2023a_deep_learning_based}
Hu, W.-C., Chen, L.-B., Hsieh, M.-H., Ting, Y.-K., 2023. A deep-learning-based
  fast counting methodology using density estimation for counting shrimp
  larvae. IEEE Sensors Journal 23~(1), 527--535.

\bibitem[{Jocher et~al.(2023)Jocher, Chaurasia, and Qiu}]{jocher2023yolo}
Jocher, G., Chaurasia, A., Qiu, J., 2023. Yolo by ultralytics. URL:
  https://github. com/ultralytics/ultralytics.

\bibitem[{Kakehi et~al.(2021)Kakehi, Sekiuchi, Ito, Ueno, Takeuchi, Suzuki, and
  Togawa}]{kakehi2021identification}
Kakehi, S., Sekiuchi, T., Ito, H., Ueno, S., Takeuchi, Y., Suzuki, K., Togawa,
  M., 2021. Identification and counting of pacific oyster crassostrea gigas
  larvae by object detection using deep learning. Aquacultural Engineering 95,
  102197.

\bibitem[{Krishna Moorthy~Babu and Jesson(2023)}]{krishna2023computer}
Krishna Moorthy~Babu, Daniel~Bentall, D. T. A. M. P. W. F. H. T. L. N. P. L. T.
  M.~W., Jesson, L.~K., 2023. Computer vision in aquaculture: a case study of
  juvenile fish counting. Journal of the Royal Society of New Zealand 53~(1),
  52--68.

\bibitem[{Li et~al.(2023)Li, Zhu, Zhang, Xu, and Li}]{li2023lightweight}
Li, W., Zhu, Q., Zhang, H., Xu, Z., Li, Z., 2023. A lightweight network for
  portable fry counting devices. Applied Soft Computing, 110140.

\bibitem[{Liu et~al.(2023)Liu, Xu, Cheng, Chen, Dou, Bi, and
  Zhao}]{liu2023shrimpseed_net}
Liu, D., Xu, B., Cheng, Y., Chen, H., Dou, Y., Bi, H., Zhao, Y., 2023.
  Shrimpseed\_net: Counting of shrimp seed using deep learning on smartphones
  for aquaculture. IEEE Access 11, 85441--85450.

\bibitem[{Lv et~al.(2023)Lv, Xu, Zhao, Wang, Wei, Cui, Du, Dang, and
  Liu}]{lv2023detrs}
Lv, W., Xu, S., Zhao, Y., Wang, G., Wei, J., Cui, C., Du, Y., Dang, Q., Liu,
  Y., 2023. Detrs beat yolos on real-time object detection.

\bibitem[{Rothschild et~al.(2023)Rothschild, Aflalo, Kedem, Farjon, Yitzhaky,
  Sagi, and Edan}]{rothschild2023computer}
Rothschild, C., Aflalo, E.~D., Kedem, I., Farjon, G., Yitzhaky, Y., Sagi, A.,
  Edan, Y., 2023. Computer vision system for counting crustacean larvae by
  detection. Smart Agricultural Technology 5, 100289.

\bibitem[{Vaswani et~al.(2017)Vaswani, Shazeer, Parmar, Uszkoreit, Jones,
  Gomez, Kaiser, and Polosukhin}]{vaswani2017attention}
Vaswani, A., Shazeer, N., Parmar, N., Uszkoreit, J., Jones, L., Gomez, A.~N.,
  Kaiser, {\L}., Polosukhin, I., 2017. Attention is all you need. Advances in
  neural information processing systems 30.

\bibitem[{Yu et~al.(2022)Yu, Wang, An, and Wei}]{yu2022counting}
Yu, X., Wang, Y., An, D., Wei, Y., 2022. Counting method for cultured fishes
  based on multi-modules and attention mechanism. Aquacultural Engineering 96,
  102215.

\bibitem[{Zhao et~al.(2021)Zhao, Zhang, Liu, Wang, Zhu, Li, and
  Zhao}]{zhao2021application}
Zhao, S., Zhang, S., Liu, J., Wang, H., Zhu, J., Li, D., Zhao, R., 2021.
  Application of machine learning in intelligent fish aquaculture: A review.
  Aquaculture 540, 736724.

\bibitem[{Zhu et~al.(2020)Zhu, Su, Lu, Li, Wang, and Dai}]{zhu2020deformable}
Zhu, X., Su, W., Lu, L., Li, B., Wang, X., Dai, J., 2020. Deformable detr:
  Deformable transformers for end-to-end object detection. arXiv preprint
  arXiv:2010.04159.

\end{thebibliography}

\end{document}